%% file: neurips_2025.tex
\documentclass{article}

\usepackage[preprint]{neurips_2026}

\usepackage[utf8]{inputenc} 
\usepackage[T1]{fontenc}    
\usepackage{hyperref}       
\usepackage{url}            
\usepackage{booktabs}       
\usepackage{amsfonts}       
\usepackage{nicefrac}       
\usepackage{microtype}      
\usepackage{xcolor}         
\usepackage{graphicx}
\usepackage{amssymb, amsfonts}
\usepackage{amsmath}
\usepackage{multirow}
\usepackage{colortbl}
\usepackage{xcolor}
\usepackage{pifont}
\usepackage[most]{tcolorbox}
\usepackage{enumitem}
\usepackage{pifont}
\usepackage{wrapfig}

\newcommand{\cmark}{\textcolor{green!60!black}{\ding{51}}}
\newcommand{\xmark}{\textcolor{gray!70}{\ding{55}}}
\definecolor{rowgray}{gray}{0.92}

\title{StreamOV: Streaming Omni-Video Understanding via Evidence-Guided Memory and Response Triggering}

\author{%
  Ming Xie$^{1,2}$\thanks{Equal contribution: mxie24@m.fudan.edu.cn} \quad
  Zizheng Huang$^{1,3}$\footnotemark[1] \quad
  Xudong Tan$^{2}$\footnotemark[1] \quad
  Chao Wang$^{2}$\footnotemark[1] \\
  \textbf{Xiangyu Zeng$^{3}$ \quad
  Wenxiao Wu$^{1,5}$ \quad
  Tao Chen$^{1,2}$\thanks{Corresponding author: yanweifu@fudan.edu.cn} \quad
  Limin Wang$^{3,4}$\footnotemark[2] \quad
  Yanwei Fu$^{1,2}$\footnotemark[2]} \\
  $^{1}$Shanghai Innovation Institute \quad
  $^{2}$Fudan University \quad
  $^{3}$Nanjing University \\
  $^{4}$Shanghai Artificial Intelligence Laboratory \quad
  $^{5}$Huazhong University of Science and Technology
}

\begin{document}

\maketitle

\begin{abstract}
While streaming omni-video understanding demands continuous perception and proactive, real-time interaction, this crucial area remains largely under-explored. Current omni-modal methods are inherently designed for offline settings, limiting their applicability in streaming scenarios due to two fundamental flaws. First, they lack robust mechanisms to manage continuously growing audio-visual context over long horizons and cannot autonomously initiate responses at opportune moments. Second, existing benchmarks are predominantly confined to offline, single-turn question answering, failing to capture continuous, multi-turn streaming interactions. To bridge these gaps, we propose StreamOV, a novel \textbf{Stream}ing \textbf{O}mni-\textbf{V}ideo understanding framework for efficient online audio-visual reasoning with bounded memory and proactive response triggering. Specifically, StreamOV introduces a multimodal evidence-guided long-short term memory that condenses historical audio-visual context into compact informative evidence under a fixed budget. It further employs a hidden-state-driven trigger to decide when to respond, avoiding explicit silence-token generation and external routers. We also curate SOVBench, the first comprehensive benchmark for online, multi-turn omni-modal evaluation. Extensive experiments show that StreamOV achieves state-of-the-art performance across diverse streaming and omni-video benchmarks, demonstrating its effectiveness for both online and offline video understanding.

\end{abstract}

\begin{figure}[t]
    \centering
    \includegraphics[width=1.0\linewidth]{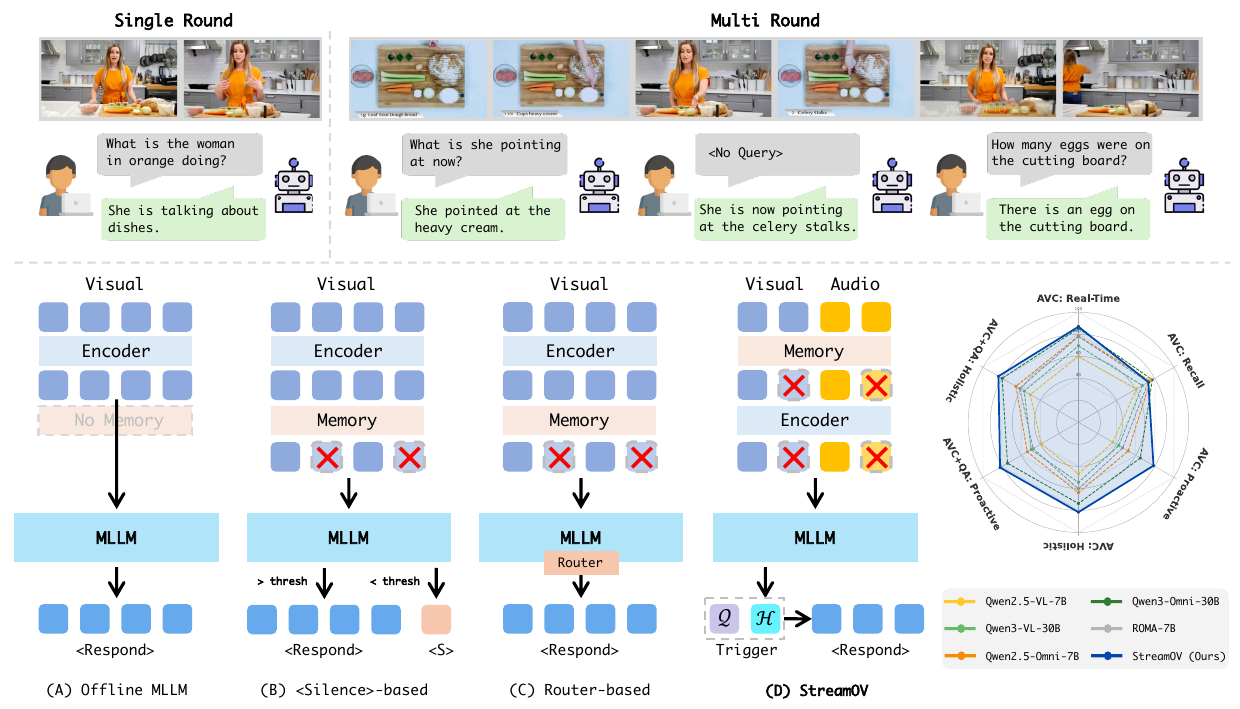}
    \caption{\textbf{Overview of StreamOV and SOVBench.} Streaming omni-video understanding requires historical multimodal memory and proactive response decisions. Compared with offline MLLMs, silence-based methods, and router-based methods, StreamOV uses compact multimodal memory and a lightweight trigger to respond efficiently without an external router. The radar chart shows state-of-the-art performance across multiple dimensions.}
    \label{fig:teaser}
    \vspace{-4pt}
\end{figure}

\section{Introduction}

Video understanding is moving from offline perception toward continuous interaction. In real-world scenarios such as long-form multimedia streams, a model cannot assume access to a complete video before answering. Instead, it must process synchronized visual and audio signals as they arrive, maintain useful historical context under bounded computation, and determine whether the current evidence is sufficient to generate a response. We refer to this setting as \textit{streaming omni-video understanding}, where the model is expected to perform online audio-visual perception, long-horizon reasoning, and proactive interaction over a continuous multimodal stream.

Despite rapid progress in large video and omni-modal models~\cite{Qwen25Omni,xu2025qwen3,bai2025qwen25vltechnicalreport,bai2025qwen3}, most existing systems remain designed for offline inference, where a fixed set of frames, audio segments, or the full video context is provided before generation. Such a formulation is fundamentally mismatched with streaming scenarios: the audio-visual context grows continuously, making naive context accumulation inefficient, and the model lacks an explicit mechanism for deciding \textit{when} to respond. Existing streaming video understanding methods partially address online perception through buffers, recurrent states, or token compression~\cite{chen2024videollm,huang2024online,di2025streaming,ning2024inf,zeng2025streamforest}, but they are predominantly visual-centric and overlook synchronized acoustic evidence. 
As shown in Fig.~\ref{fig:teaser}, existing proactive streaming methods usually rely on silence-token generation~\cite{chen2024videollm,xia2025streaming,ding2025streammind} or external routers~\cite{qian2025dispider,azad2026streamready}. Silence tokens introduce non-semantic control outputs that may perturb the pretrained generation distribution, while routers delegate response decisions to auxiliary models with limited multimodal reasoning.

Evaluation also remains insufficient. Existing video benchmarks mainly focus on offline or single-turn question answering, while current streaming benchmarks either emphasize visual-only perception or evaluate audio-visual QA as independent single-round tasks. They therefore fail to measure whether a model can preserve long-range multimodal context, handle multi-round temporal interactions, infer latent user intent from dialogue history, and intentionally remain silent when queried evidence is absent. To bridge these gaps, we propose \textbf{StreamOV}, a streaming omni-video understanding framework built on an omni-modal large language model. StreamOV constructs multimodal evidence from both query-agnostic stream dynamics and query-aware semantic relevance, routes it into visual-only, audio-only, and audio-visual-aligned cues, and maintains a long-short term memory that preserves dense recent observations while sparsely retaining informative historical evidence under a fixed budget. For proactive interaction, StreamOV further introduces an MLLM-as-a-trigger mechanism that probes the hidden states of early decoding steps to predict whether to \textit{Respond} or \textit{Wait}, avoiding both explicit \texttt{<silence>} token generation and external routers.

We further construct \textbf{SOVBench}, the first benchmark dedicated to online multi-round omni-video evaluation. SOVBench-O evaluates continuous audio-visual comprehension across Real-Time, Recall, and Proactive interaction paradigms, while SOVBench-T directly evaluates operational proactivity as a response-triggering task, requiring models to respond when queried evidence appears and remain silent when it is absent. Extensive experiments on SOVBench, StreamingBench, OVO-Bench, and offline audio-visual benchmarks demonstrate that StreamOV achieves strong performance across streaming omni-modal, visual-only, and offline settings.

Our contributions are summarized as follows:

\begin{itemize}[leftmargin=*]

    \item We formulate \textit{streaming omni-video understanding} as an online audio-visual interaction problem requiring bounded memory, long-horizon reasoning, and proactive response timing, and construct \textbf{SOVBench}, the first benchmark for online multi-round omni-video evaluation with both comprehension and triggering tasks.

    \item We propose \textbf{StreamOV}, which performs multimodal evidence routing and long-short term memory to efficiently preserve informative visual, audio, and audio-visual cues under streaming setting.

    \item We introduce a hidden-state-driven response trigger that leverages the MLLM's early decoding states to decide whether to respond or wait, avoiding explicit silence-token generation and external routers while achieving state-of-the-art or competitive performance across multiple benchmarks.

\end{itemize}

\section{Related Works}

\textbf{Offline Video Understanding. }Large Vision-Language Models (LVLMs) have driven rapid progress in offline video understanding. Early works (e.g., VideoChat~\cite{li2025videochat}, Video-LLaVA~\cite{lin2024video}, LLaVA-NeXT-Video~\cite{zhang2024llavanextvideo}) adapted image-based LVLMs via sparse frame sampling and spatial-temporal modeling. This visual-centric approach recently culminated in models like Qwen-VL series~\cite{Qwen-VL,wang2024qwen2,bai2025qwen25vltechnicalreport,bai2025qwen3}, which excel at fine-grained spatial-temporal reasoning over extended context windows. However, recognizing video's multi-sensory nature, the field is now shifting toward omni-modal understanding. Moving beyond simplistic late-fusion approaches, unified end-to-end architectures like Qwen-Omni series~\cite{Qwen25Omni,xu2025qwen3} natively interleave audio, vision, and text. This enables joint cross-modal reasoning to accurately capture the intricate correlations between visual actions and acoustic events.


\textbf{Streaming Omni-Video Understanding.}
Compared with offline video reasoning and understanding, streaming video understanding requires low latency and efficient memory management~\cite{xia2025streaming}. Early works~\cite{chen2024videollm, huang2024online} established this paradigm with streaming buffers, while subsequent studies improve long-context modeling through KV cache compression and recurrent memory~\cite{di2025streaming, ning2024inf}. Computational efficiency has been further optimized via adaptive resource allocation~\cite{wu2024videollm} and hierarchical token processing~\cite{zhang2025flash, zeng2025streamforest, yao2025timechat}. For interactive scenarios, recent models~\cite{liu2024streamchat, ning2025livevlm, chen2025livecc} enable real-time dialogue and continuous perception. However, most existing streaming video understanding methods still focus primarily on the visual modality, leaving synchronized acoustic streams underexplored~\cite{tian2026roma}. In this work, we study Streaming Omni-Video Understanding, where models must continuously perceive, reason, and interact over synchronized visual and audio streams under an online setting.

\textbf{Online Video Benchmarks. }MovieChat-1K~\cite{song2024moviechat} and VideoLLM-online~\cite{chen2024videollm} pioneered long-term memory and streaming evaluation protocols, while OVOBench~\cite{niu2025ovo} and VStream-QA ~\cite{zhang2025flash} introduced objective metrics to ensure causal consistency in real-time perception. More recently, StreamingBench~\cite{lin2024streamingbench} and SVBench~\cite{xiong2025streaming} established comprehensive benchmarks for streaming reasoning and multi-round interactions. However, while StreamingBench incorporates audio-visual QA, it is limited to independent single-round tasks; SVBench, on the other hand, supports multi-round interactions but lacks comprehensive audio-visual grounding. Consequently, there remains a critical gap in evaluating a model's multi-round online audio-visual QA capabilities.

\section{SOVBench}\label{sec:sov}

\begin{figure}
    \centering
    \includegraphics[width=\linewidth]{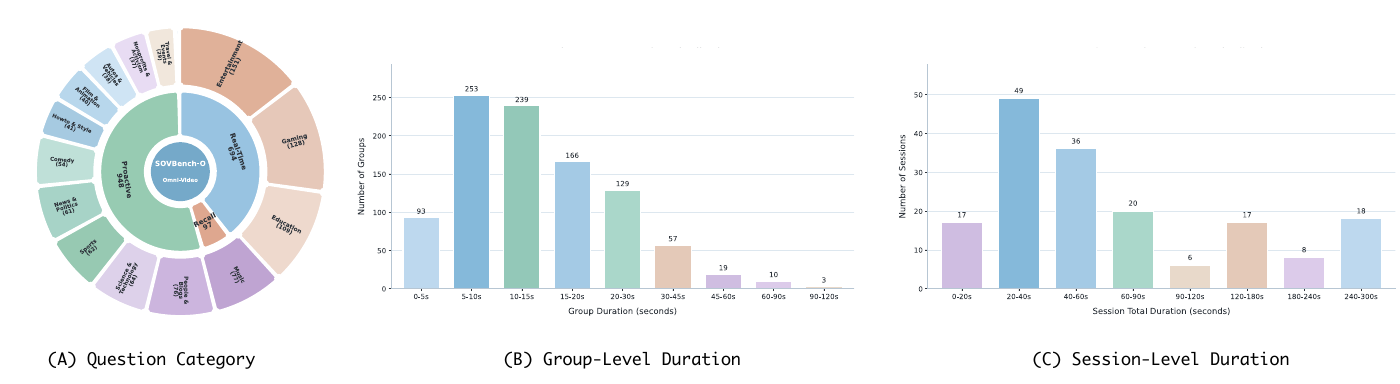}
    \caption{Statistics of SOVBench-O. (a) Distribution of question categories across different real-world domains. (b) Distribution of group-level (QAs with follow-up reasoning) durations, where each group corresponds to a temporally coherent local interaction segment. (c) Distribution of session-level durations, reflecting the temporal span of complete streaming sessions.}\label{fig:data_statis}\vspace{-6pt}
\end{figure}

To comprehensively evaluate streaming omni-video understanding, we introduce SOVBench, which consists of two complementary components. SOVBench-O evaluates continuous comprehension accuracy through multi-round temporal interactions, while SOVBench-T assesses the model's proactive response decision-making and intentional silence capabilities.

\subsection{Data Source Filtering \& Generation}\label{sec:data_filter_gen}
To construct a robust benchmark for streaming omni-video understanding (SOVU), we select FineVideo~\cite{Farré2024FineVideo} as our primary data source. FineVideo provides large-scale internet videos paired with exceptionally dense metadata, including global descriptions, scene-level narratives, localized activities, prop interactions, and timestamped Automatic Speech Recognition transcripts. These fine-grained, multimodal annotations serve as an ideal foundation for evaluating continuous comprehension in a streaming context. While FineVideo offers extensive annotations, the complexity of SOVU necessitates data with dense information, and strong cross-modal alignment. To ensure the highest quality of our benchmark and eliminate noisy or trivial samples, we employ an advanced large language model~\cite{bai2025qwen3} as an automated quality assessor to filter the dataset based on its raw metadata. We design a rigorous evaluation prompt that scores each video on a scale of $0$ to $10$ across five critical dimensions: Visual Dynamism, Narrative Coherence, Information Density, Audio-Visual Alignment and Reasoning Value (see Supp.~\ref{supp:data_filtering} for details). Alongside the quantitative scores, the assessor generates detailed rationales. Finally, we apply a strict score threshold for filtering and conduct manual sanity checks on these rationales to guarantee the reliability of the selection process. To subsequently generate high-quality multi-round QAs, we extract the fine-grained multimodal metadata from the filtered videos and reorganize it into a strictly chronological timeline integrating visual events, acoustic signals, and transcripts. We then prompt Gemini~\cite{comanici2025gemini} with this temporal blueprint to formulate multi-round interactions that are strictly grounded in the timeline while covering short-term occurrences, long-term historical dependencies, and cross-modal reasoning. The generated QAs are manually verified by annotators to ensure temporal correctness and cross-modal grounding.

\subsection{SOVBench-O: Multi-round Omni QA Benchmark}\label{sec:sovbencho}
To comprehensively evaluate audio-visual capabilities, we construct SOVBench-O, a benchmark containing multi-round QAs targeting fine-grained visual details (\textit{e.g.}, scenes, appearances, expressions, text) and key auditory elements (\textit{e.g.}, acoustic events, speech). While some questions assess single-modality capabilities, a subset of questions requires joint multimodal reasoning. To strictly evaluate continuous streaming comprehension, we design three temporal interaction paradigms: \textbf{Real-Time}, \textbf{Recall}, and \textbf{Proactive}. Real-Time QA tests on-the-fly understanding of immediate, ongoing events. Recall QA tests long-term memory retention by querying information that appeared earlier in the video stream.
Finally, Proactive QA assesses autonomous contextual awareness across two distinct dimensions: \textit{Cognitive Proactivity} and \textit{Operational Proactivity}. Within SOVBench-O, we specifically evaluate cognitive proactivity by testing implicit context understanding. In such scenarios, explicit questions are entirely omitted and only candidate options are provided if the current turn is highly correlated with the previous round, forcing the model to infer the latent intent of the user from short-term dialogue history. Through these diverse paradigms, SOVBench-O provides a rigorous evaluation of cognitive comprehension and contextual reasoning in continuous streams.

\subsection{SOVBench-T: Omni Response Triggering Benchmark}\label{sec:sovbencht}
Beyond content accuracy, the streaming nature of SOVU demands exceptional operational proactivity (optimal triggering  judgment), since knowing whether to respond is as critical as what to say. Existing approaches often incorporate specialized tokens, such as \texttt{<silence>}~\cite{xia2025streaming,chen2024videollm}, into the training process to endow the model with the capacity for proactive response and intentional silence. However, their evaluations typically rely on auxiliary tasks~\cite{qian2025dispider,zheng2026garde}, such as temporal grounding, to indirectly reflect the effectiveness of these mechanisms. To provide a more intuitive evaluation of a model's proactive response capability in a video stream, we construct SOVBench-T, a benchmark centered on a binary classification task to evaluate this fundamental ability of online models. Specifically, positive samples consist of content where the queried omni-information eventually appears, requiring the model to autonomously trigger a timely response within a temporal window. In contrast, negative samples involve queries for information that is entirely absent, forcing the model to demonstrate its capacity for intentional silence and refrain from generating hallucinated responses. By focusing on this binary verification paradigm, SOVBench-T provides a direct measure of whether a model's proactive behavior is grounded in actual omni-perceptions.

\subsection{Benchmark Statistics}
As shown in Fig.~\ref{fig:data_statis}, after filtering and manual verification, SOVBench-O contains 172 streaming sessions from 172 distinct FineVideo videos, comprising 1,739 question-answer turns organized into 969 temporally coherent dialogue groups. Each group corresponds to a short local interaction segment, while successive groups within the same session evaluate the model's ability to preserve long-range context over time. These groups cover both single-turn and multi-turn interactions, with many requiring follow-up reasoning rather than isolated one-shot recognition. On average, each group contains 1.79 turns and spans 15.46 seconds of video content, with a median duration of 12.08 seconds. Among all groups, 368 are single-turn interactions, while 601 contain multiple turns, indicating that a substantial portion of the benchmark evaluates follow-up reasoning. Beyond temporal diversity, SOVBench-O spans 15 top-level categories and 86 fine-grained semantic categories across diverse real-world domains, preventing the evaluation from collapsing into narrow action recognition and requiring models to reason over varied audio-visual evidence, narration styles, and conversational intents. SOVBench-T further focuses on response triggering in real-world streaming scenarios. It contains 226 valid samples, with 120 positive samples requiring a response and 106 negative samples where the model should remain silent, forming a near-balanced binary evaluation set. More statistics are provided in Supp.~\ref{supp:bench_statis}.

\section{Methodology}

\textbf{Task Formulation. }In contrast to offline video understanding, which assumes access to a complete, fixed-length video sequence, streaming omni-video understanding operates on a continuous and potentially infinite multi-modal stream. We denote this stream as $\{x_1, x_2, \dots, x_t, \dots\}$, where each observation $x_t = \{v_t, a_t\}$ at time step $t$ consists of visual frames $v_t$ and synchronized audio segments $a_t$. The objective of online model $\phi$ is to perform online perception and interaction, generating a response $y_t$  based on the incoming stream and a task-specific query $q_t$.

Due to the potential infinite horizon of streams and memory constraints, it is computationally prohibitive to retain the entire history of observations $\{x_\tau\}_{\tau=1}^t$ as input. To address this, we formulate the streaming process as a state transition. We define a memory state $M_t$ to represent the historical context up to time $t$. This memory state is updated recursively by integrating the new observation $x_t$ into the previous state $M_{t-1}$, and the output $y_t$ is generated based on the updated memory:
\begin{equation}
    M_t = \mathcal{F}_{mem}(M_{t-1}, x_t), \quad y_t = \phi(M_t, q_t),
\end{equation}
where $\mathcal{F}_{mem}$ denotes the memory update function. Here, $M_t$ serves as a compact, bounded-size summary of the accumulated stream, flexibly taking arbitrary forms such as token sequences, keyframe buffers, or other representations.
Beyond information compression, a critical challenge in streaming setting is determining the optimal timing for interaction. Unlike offline systems, a streaming model must autonomously decide whether to remain silent or to respond proactively based on $M_t$. This necessitates identifying opportune moments when accumulated evidence is sufficient.

\begin{figure*}
    \centering
    \includegraphics[width=1.0\linewidth]{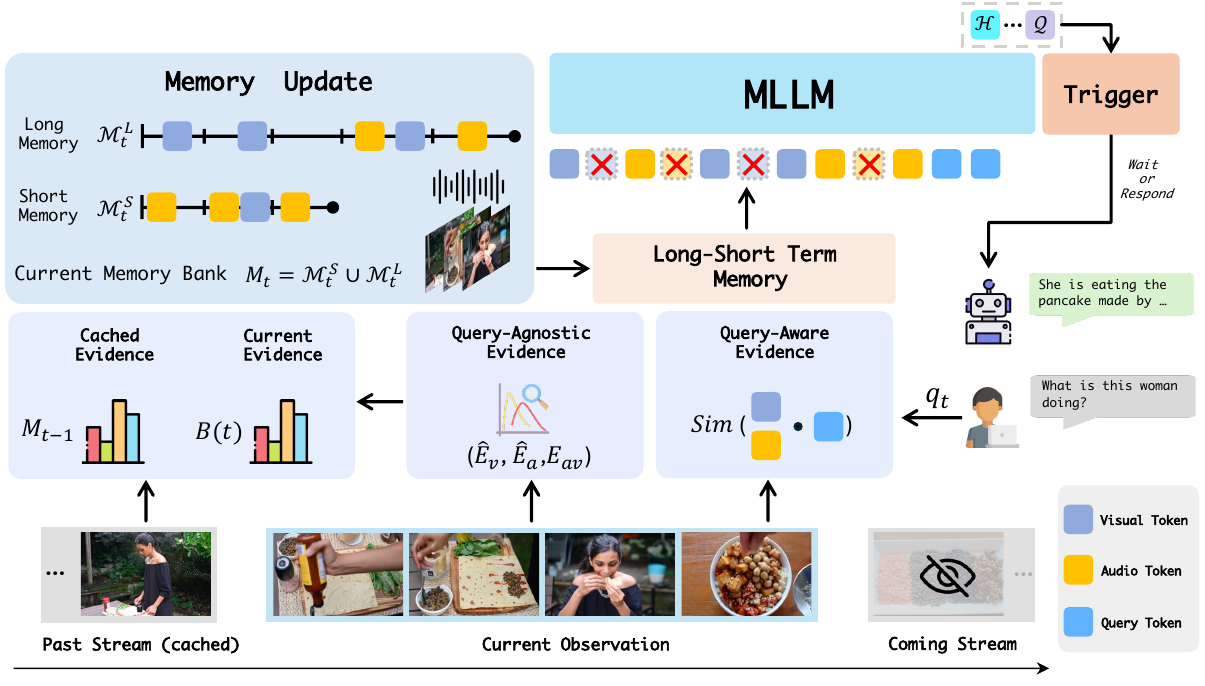}
    \caption{Framework overview of \textbf{StreamOV}. Given a omni-video stream, StreamOV constructs multimodal evidence to update a long-short term memory. The updated memory is fed into the frozen MLLM, and a lightweight hidden-state trigger decides whether to wait or generate a response.}\label{fig:framework}
    \vspace{-4pt}
\end{figure*}

\subsection{Multimodal Evidence Construction}\label{sec:av_evidence}

To mitigate the redundancy of continuous video streaming, we propose a dual-perspective evidence construction mechanism that quantifies both query-agnostic stream dynamics and query-aware semantic relevance, allowing the online system to persist the most informative segments under a limited memory budget. We illustrate the construction process in Fig.~\ref{fig:framework}.

We first extract query-agnostic characteristics to capture intrinsic stream dynamics. These include visual change $S_v$ computed by consecutive frame differencing, audio saliency $S_a$ derived from waveform peak detection, and audio-visual co-burst $S_{cob}$ indicating synchronized multimodal events. Concurrently, we evaluate query-aware semantics by measuring the relevance between the query $q_t$ and the multimodal stream. Specifically, we leverage pre-trained encoder~\cite{radford2021learning,elizalde2023clap} to obtain the visual semantic score $S_{qv}$ and audio semantic score $S_{qa}$, respectively. Since these metrics have different numerical ranges, we calculate the normalized value $S = r / r_{max}$ within current observation before aggregation, where $r$ denotes the zero-indexed rank of the original observation.

The normalized metrics are then integrated to identify salient segments for downstream reasoning. Since audio semantic matching is often susceptible to background noise, we introduce a gating mechanism to refine the audio evidence. The effective audio semantic score is defined as $\hat{S}_{qa}=S_{qa}\cdot\max(S_a,S_{cob})$, which prioritizes audio-query relevance only when the segment contains physically salient audio or synchronized audio-visual changes. Based on these refined metrics, we derive preliminary visual and audio evidence scores:
\begin{equation}
    E_v = \max(S_{qv}, S_v), \quad E_a = \max(\hat{S}_{qa}, S_a).
\end{equation}

To capture multimodal events, we further construct audio-visual aligned evidence $E_{av}$ from two complementary cues: semantic consistency between visual and audio relevance, and event-level synchronization across modalities. Concretely, $E_{av}$ is defined as the maximum of the semantic alignment $\min(S_{qv},\hat{S}_{qa})$ and the event-driven burst $\min(E_v,E_a,S_{cob})$. To avoid assigning the same synchronized evidence redundantly to both modality-specific branches, we decouple the aligned evidence from the visual and audio scores:
\begin{equation}
    \hat{E}_v = [E_v - E_{av}]_+, \quad \hat{E}_a = [E_a - E_{av}]_+,
\end{equation}
where $[\cdot]_+=\max(\cdot,0)$. The final evidence is given by $\max(\hat{E}_v,\hat{E}_a,E_{av})$, and the temporal window is routed to three disentangled evidence, namely visual-only, audio-only, or audio-visual-aligned. This routing strategy provides a compact and structured evidence representation for streaming reasoning.

\subsection{Long-Short Term Memory Update}\label{sec:memory_update}

Based on the constructed evidence, we update the streaming memory with a long-short term strategy. The short-term memory preserves dense observations within the current window. The long-term memory stores a sparse set of informative observations selected from both current evidence and cached historical evidence.

For each observation $x_t$, we determine its base evidence score $B(t)$ as the maximum value among the three routed evidence types:
\begin{equation}
    B(t)=\max(\hat{E}_v(t),\hat{E}_a(t),E_{av}(t)).
\end{equation}
Since $B(t)$ is derived from the rank-normalized metrics introduced in Sec. \ref{sec:av_evidence}, it functions as a \textit{local importance score} that quantifies how much an observation stands out within its own temporal neighborhood. This normalization effectively implements a local importance sampling mechanism, ensuring that salient multimodal events are identified and compared under a unified relative scale.

The streaming memory $M_t$ is structured into a short-term memory $\mathcal{M}^S_t$ and a long-term memory $\mathcal{M}^L_t$, both of which are populated via score-based selection but differ in their temporal density.
The short-term memory $\mathcal{M}^S_t$ retains the dense Top-$K_S$ observations from the current temporal window, due to the limited duration.
In contrast, the long-term memory $\mathcal{M}^L_t$ maintains a fixed Top-$K_L$ budget of informative observations curated from the entire historical stream. As the stream progresses, $\mathcal{M}^L_t$ is dynamically updated by evaluating a candidate pool composed of the existing long-term entries and the outgoing observations from the short-term window. This results in a sparse but globally significant set of multi-modal events.

The final memory state is constructed by the union of the two buffers: $M_t = \mathcal{M}^S_t \cup \mathcal{M}^L_t$. Observations that appear in both buffers are deduplicated. All retained observations are then serialized in chronological order and fed into the base model in an interleaved visual-audio format. This memory design preserves fine-grained recent context while enabling the model to perform long-range reasoning over a curated history of salient multi-modal evidence.



\subsection{Response Trigger}\label{sec:trigger}
To achieve operational proactivity, existing approaches typically rely on explicit auto-regressive generation of specialized action tokens~\cite{chen2024videollm,xia2025streaming,ding2025streammind} or employ cascade models~\cite{qian2025dispider,azad2026streamready}. However, the former forces a Large Language Model to explicitly generate non-semantic silence tokens, which inevitably disrupts its pre-trained distribution, while the latter delegates triggering decisions to auxiliary small models. These small routers inherently lack the deep reasoning capacity required for complex queries and fall short of the primary MLLM's comprehension, especially given the current scarcity of reliable small-scale omni-modal models. 

Instead of relying on inferior external routers, we propose regarding MLLM itself as a trigger, leveraging a more intrinsic and efficient mechanism: the model's intentional readiness to respond is already encoded in the hidden states of initial decoding steps. Based on this motivation, we propose a hidden state-driven trigger that directly probes the model's internal cognitive state to make autonomous response decisions. Specifically, at any timestep $t$, when the model receives the accumulated memory $M_t$ and query $q_t$, it initiates the prefilling and decoding process. 
Let $\mathcal{H}_t = \{h_{t,0}, h_{t,1}, \dots, h_{t,k}\}$ denote the hidden states extracted from the last decoder layer, where $h_{t,0}$ is the hidden state at the final input position after prefilling, and $h_{t,i}$ for $i \geq 1$ is the hidden state at the newly generated position during the $i$-th decoding step. Each $h_{t,i} \in \mathbb{R}^D$ serves as a semantic latent representation of the model's current response intent.
Rather than generating the full response, we utilize this leading prefix of hidden states as a compact proxy for the model's confidence and intent. To aggregate this internal cognitive evidence, we introduce a lightweight Cross-Attention Trigger module, parameterized by $\theta_{tr}$. This module employs a learnable query vector $Q_{tr} \in \mathbb{R}^{1 \times D}$ to attend to the sequence of prefix hidden states $\mathcal{H}_t$, which serve as keys and values:
\begin{equation}
    z_t = \text{Softmax}\left(\frac{Q_{tr} \mathcal{H}_t^T}{\sqrt{D}}\right) \mathcal{H}_t, \quad p_t = \text{MLP}(z_t; \theta_{tr}),
\end{equation}
where $z_t \in \mathbb{R}^D$ is the context-aware response intent representation, which is then fed into a classification head to yield the binary triggering logits $p_t \in \mathbb{R}^2$. The prediction $p_t$ categorizes the current state into two actions: \textit{Respond} or \textit{Wait}. If the accumulated multimodal evidence in $M_t$ is insufficient to address $q_t$, the trigger outputs \textit{Wait}, and the generation process is seamlessly truncated to preserve computational resources. Conversely, if \textit{Respond} is triggered, the model proceeds with full auto-regressive generation to yield $y_t$. During training, the trigger takes the proposed memory representation as input and we optimize the trigger $\theta_{tr}$ using a cross-entropy loss with MLLM keeping frozen. In our implementation, we only use $h_{t,0}$ for trigger prediction, which allows the triggering decision to be made immediately after prefilling and before any auto-regressive decoding step.

\section{Experiments}\label{sec:exp}


\textbf{Baseline and Benchmarks. }
We evaluate our method against representative Qwen-series models, including Qwen-VL~\cite{Qwen-VL,bai2025qwen25vltechnicalreport} models at both 7B and 30B scales and Qwen-Omni~\cite{Qwen25Omni,xu2025qwen3} models at both 7B and 30B scales. Since Qwen models are offline models, we further compare with state-of-the-art online streaming models to provide a comprehensive evaluation under streaming scenarios. SOVBench provides two evaluation settings. The first setting directly takes the audio-visual stream as input, while the second further accumulates previous QA pairs as dialogue context for multi-round evaluation. Besides, we adopt StreamingBench~\cite{lin2024streamingbench} and OVO-Bench~\cite{niu2025ovo} as additional benchmarks for audio-visual QA and visual-only QA, respectively. We also report results on two offline audio-visual QA benchmarks, Video-Holmes~\cite{cheng2025video} and Daily-Omni~\cite{zhou2025daily}, to demonstrate that our method achieves superior performance beyond streaming-specific evaluation.

\textbf{Implementation Details. }
Our method is built upon Qwen3-Omni-30B-A3B, which contains 30B total parameters with 3B activated parameters during inference. We keep the backbone model frozen and train only a lightweight trigger module with 18.9M trainable parameters. The trigger is optimized using AdamW optimizer with a learning rate of $3 \times 10^{-4}$ and a batch size of 32. The trigger training data refers to Supp.~\ref{supp:triggers_traindata} for details. We only use \textbf{one hidden state} for trigger training.
During inference, the trained trigger determines whether the model should respond or remain silent at each streaming step. 
We set the streaming budget to 64 frames in SOVBench and 32 frames in others. All baseline methods are sampled at 1 FPS. All experiments are conducted on 80G 8 $\times$ H100.

\textbf{Metrics. }Besides content accuracy on different benchmarks, we report category-wise accuracy for Real-Time, Recall, and Proactive questions on SOVBench-O. We also evaluate the silence mechanism of online models on SOVBench-T, formulating it as a binary classification task and reporting precision, recall, and F1 score to measure the model's proactive capability.

\subsection{Evaluation on Audio-Visual Benchmarks}

Both our SOVBench and StreamingBench provide multimodal question-answering evaluations. As shown in Tab.~\ref{tab:sovbench}, SOVBench is the first multimodal multi-round QA benchmark designed for streaming omni-video understanding. The results show that SOVBench poses substantial challenges to existing omni-modal models: even Qwen3-Omni-30B achieves only 73.7\% and 79.9\%, whereas our StreamOV improves the performance to 81.6\% and 83.8\%, corresponding to a gain of 7.9\% and 3.9\%, respectively.
Although Qwen3-Omni performs better in some scenarios, it benefits from a less constrained offline setting. Our method targets budgeted online inference and achieves superior proactive performance, a key metric in streaming settings.
Furthermore, as reported in Tab.~\ref{tab:streamingbench}, StreamOV consistently outperforms audio-enabled Qwen3-Omni (+7.6\%) and ROMA~\cite{tian2026roma}(+22.5\%) on StreamingBench omni subset, achieving higher quantitative metrics in Audio-Visual QA setting.

On offline multi-modal benchmarks, as shown in Tab.~\ref{tab:offline_av}, StreamOV achieves competitive results on the reasoning-oriented Video-Holmes benchmark. Moreover, StreamOV consistently obtains superior performance on Daily-Omni. We attribute these improvements to our effective evidence routing strategy, which enables the model to better identify and preserve informative audio-visual cues.

\begin{table}[t]
\centering
\caption{Evaluation on proposed SOVBench. Offline models use 1 FPS for input construction (see duration statistics in Fig.~\ref{fig:data_statis}), whereas StreamOV sets an online budget of 64 frames.}\label{tab:sovbench}
\resizebox{\textwidth}{!}{%
\begin{tabular}{lcc|ccc|c|ccc|c|ccc} 
\toprule
\multirow{4}{*}{\textbf{Model}} & \multirow{4}{*}{\textbf{Audio}} & \multirow{4}{*}{\textbf{frames}} & \multicolumn{11}{c}{\textbf{SOVBench}} \\ 
\cmidrule(l){4-14}
& & & \multicolumn{4}{c|}{\textbf{Audio-Visual Context}} 
  & \multicolumn{4}{c|}{\textbf{Audio-Visual Context \& QA context}} 
  & \multicolumn{3}{c}{\textbf{SOVBench-T}} \\
\cmidrule(lr){4-7} \cmidrule(lr){8-11} \cmidrule(l){12-14}
& & & Real-Time & Recall & Proactive & Avg.
    & Real-Time & Recall & Proactive & Avg.
    & P/R-1 & P/R-0 & F1 \\
\midrule
\rowcolor{rowgray}
\multicolumn{14}{c}{\textbf{Closed-source Multimodal Models}} \\
\midrule
Gemini 2.5 Flash & \cmark & - 
& 87.0 & 80.4 & 66.8 & 75.6  
& 88.9 & 85.6 & 70.8 & 78.8  
& \multicolumn{3}{|c}{\multirow{2}{*}{\shortstack{\textit{Offline} \\ \textit{Models}}}} \\
Gemini 3.0 Flash & \cmark & - 
& 89.5 & 87.6 & 82.6 & 85.6  
& 90.3 & 89.7 & 87.2 & 88.6  
& \multicolumn{3}{|c}{} \\
\midrule
\rowcolor{rowgray}
\multicolumn{14}{c}{\textbf{Open-source Offline VideoLLMs}} \\
\midrule
Qwen2.5-VL-7B & \xmark & 1 fps 
& 59.7 & 60.8 & 36.0 & 46.8  
& 63.8 & 70.1 & 38.9 & 50.6  
& \multicolumn{3}{|c}{\multirow{4}{*}{\shortstack{\textit{Offline} \\ \textit{Models}}}} \\
Qwen2.5-Omni-7B & \cmark & 1 fps 
& 78.4 & 75.3 & 52.1 & 63.9  
& 79.5 & 81.4 & 53.6 & 65.5  
& \multicolumn{3}{|c}{} \\
Qwen3-VL-30B-A3B & \xmark & 1 fps 
& 69.5 & 67.0 & 42.3 & 54.5  
& 65.0 & 70.1 & 49.5 & 56.8  
& \multicolumn{3}{|c}{} \\
Qwen3-Omni-30B-A3B & \cmark & 1 fps 
& 85.3 & \textbf{77.3} & 64.8 & 73.7  
& \textbf{87.3} & \textbf{82.5} & 74.3 & 79.9  
& \multicolumn{3}{|c}{} \\
\midrule
\rowcolor{rowgray}
\multicolumn{14}{c}{\textbf{Open-source Online VideoLLMs}} \\
\midrule
ROMA-7B & \cmark & 1 fps 
& 78.0 & 73.2 & 46.3 & 60.4  
& 78.1  & 79.4 & 47.5 & 61.5  
& 54.6/59.2 & 49.0/44.3 & 52.2 \\
\textbf{StreamOV} & \cmark & 64 
& \textbf{86.9} & 73.2 & \textbf{78.6} & \textbf{81.6}  
& 86.7 & 80.4 & \textbf{82.1} & \textbf{83.8}  
& \textbf{86.1/98.3} & \textbf{97.8/82.1} & \textbf{90.5} \\
\bottomrule
\end{tabular}%
}
\end{table}

\begin{table}[htb]
\centering
\caption{Evaluation on online Audio-Visual and Visual-Only benchmark: StreamingBench.}\label{tab:streamingbench}
\resizebox{\textwidth}{!}{%
\begin{tabular}{lc|ccccc|ccccccccccc}
\toprule
\multirow{4}{*}{\textbf{Model}} & \multirow{4}{*}{\textbf{Audio}} & \multicolumn{16}{c}{\textbf{StreamingBench}} \\ 
\cmidrule(l){3-18}
& & \multicolumn{5}{c|}{\textbf{Audio-Visual QA}} & \multicolumn{11}{c}{\textbf{Visual-Only QA}} \\
\cmidrule(lr){3-7} \cmidrule(l){8-18}
&  & ER & SCU & SD & MA & Avg. & OP & CR & CS & ATP & EU & TR & PR & SU & ACP & CT & Avg. \\
\midrule
\rowcolor{rowgray}
\multicolumn{18}{c}{\textbf{Open-source Offline VideoLLMs}} \\
\midrule
Qwen2.5-VL-7B & \xmark & 43.6 & 23.2 & 45.2 & 54.0 & 41.5 & 79.0 & 78.1 & 83.9 & 82.7 & 74.5 & 79.8 & 82.4 & 67.1 & 67.4 & 44.0 & 74.5 \\
Qwen2.5-Omni-7B & \cmark & 48.0 & 22.4 & 50.0 & 70.0 & 49.6 & 80.4 & 79.7 & 81.4 & 80.7 & 76.4 & 81.9 & 83.3 & 68.3 & 66.0 & 43.5 & 74.5 \\
Qwen3-VL-30B-A3B & \xmark & 44.8 & 30.4 & 46.4 & 62.4 & 46.0 & 83.7 & 80.5 & 91.2 & 87.3 & 78.3 & 83.2 & 86.1 & 69.9 & 69.1 & 44.0 & 78.1 \\
Qwen3-Omni-30B-A3B & \cmark & 53.6 & 40.8 & 70.4 & 79.2 & 61.0 & 81.5 & 81.3 & 89.9 & 85.0 & 77.6 & 83.8 & 87.0 & 69.5 & 73.7 & 42.0 & 77.9 \\
\midrule
\rowcolor{rowgray}
\multicolumn{18}{c}{\textbf{Open-source Online VideoLLMs}} \\
\midrule
Flash-VStream-7B & \xmark & \multicolumn{5}{c|}{\multirow{5}{*}{\shortstack{\textit{Visual-Only Online Models} \\ \textit{Random Guessing}}}} & 25.9 & 43.6 & 24.9 & 23.9 & 27.3 & 13.1 & 18.5 & 25.2 & 23.9 & 48.7 & 23.2 \\
VideoLLM-online-8B & \xmark & \multicolumn{5}{c|}{} & 39.1 & 40.1 & 34.5 & 31.1 & 46.0 & 32.4 & 31.5 & 34.2 & 42.5 & 27.9 & 36.0 \\
Dispider-7B & \xmark & \multicolumn{5}{c|}{} & 74.9 & 75.5 & 74.1 & 73.1 & 74.4 & 59.9 & 76.1 & 62.9 & 62.2 & 45.8 & 67.6 \\
TimeChat-Online-7B & \xmark & \multicolumn{5}{c|}{} & 80.2 & 82.0 & 79.5 & 83.3 & 76.1 & 78.5 & 78.7 & 64.6 & 69.6 & \textbf{58.0} & 75.4 \\
StreamForest-7B & \xmark & \multicolumn{5}{c|}{} & 83.1 & 82.8 & 82.7 & 84.3 & 77.5 & 78.2 & 76.9 & 69.1 & 75.6 & 54.4 & 77.3 \\
ROMA-7B & \cmark & 40.4 & 34.8 & 50.4 & 58.8 & 46.1 & 77.0 & 78.1 & 77.9 & 82.1 & 74.8 & 72.9 & 82.4 & 61.8 & 65.9 & 51.1 & 72.4 \\
\midrule
\textbf{StreamOV} & \cmark & \textbf{63.6} & \textbf{59.2} & \textbf{79.2} & \textbf{90.4} & \textbf{68.6} & \textbf{92.6} & \textbf{87.5} & \textbf{95.0} & \textbf{90.5} & \textbf{85.7} & \textbf{94.7} & \textbf{88.0} & \textbf{80.0} & \textbf{86.1} & 44.6 & \textbf{86.2} \\
\bottomrule
\end{tabular}%
}
\end{table}

\begin{table}
\centering
\caption{Evaluation on offline Audio-Visual QA benchmarks: Video Holmes and Daily-Omni.}\label{tab:offline_av}
\resizebox{\textwidth}{!}{%
\begin{tabular}{lc|cccccccc|ccccccccc}
\toprule
\multirow{2.5}{*}{\textbf{Model}} & \multirow{2.5}{*}{\textbf{Audio}} & \multicolumn{8}{c|}{\textbf{Video Holmes (32 frames)}} & \multicolumn{9}{c}{\textbf{Daily-Omni (1 FPS)}} \\
\cmidrule(lr){3-10} \cmidrule(lr){11-19}
 & & SR & IMC & TCI & TA & MHR & PAR & CTI & Avg. & AVA & CP & CU & ES & IF & RS & 30 & 60 & Avg. \\
\midrule
\rowcolor{rowgray}
\multicolumn{19}{c}{\textbf{Closed-source Multimodal Models}} \\
\midrule
Gemini 2.0 Flash & \cmark & 56.5 & 54.2 & 43.4 & 44.5 & 43.9 & 55.1 & 50.1 & 49.5 & 62.2 & 73.3 & 63.7 & 63.7 & 76.6 & 75.4 & 67.2 & 68.6 & 67.8 \\
Gemini 2.5 Flash & \cmark & 43.4 & 46.9 & 43.1 & 51.0 & 37.9 & 43.6 & 39.3 & 43.1 & 73.8 & 66.4 & 72.0 & 68.0 & 78.7 & 81.9 & 69.9 & 77.1 & 73.1 \\
GPT-4o & \xmark & 50.0 & 49.6 & 38.8 & 30.0 & 44.0 & 39.2 & 37.0 & 42.0 & 47.9 & 62.6 & 52.3 & 52.6 & 66.2 & 66.3 & 55.6 & 57.5 & 56.5 \\
\midrule
\rowcolor{rowgray}
\multicolumn{19}{c}{\textbf{Open-source Offline VideoLLMs}} \\
\midrule
Qwen2.5-VL-7B & \xmark & 34.6 & 21.4 & 15.4 & 12.5 & 17.2 & 12.4 & 15.2 & 19.0 & 37.0 & 46.6 & 33.7 & 37.9 & 52.0 & 44.0 & 39.3 & 42.4 & 40.7 \\
Qwen2.5-Omni-7B & \cmark & 41.8 & 35.1 & 23.4 & 25.0 & 28.9 & 17.5 & 25.9 & 29.0 & 34.5 & 58.8 & 47.7 & 49.7 & 63.0 & 54.9 & 48.7 & 51.1 & 49.8 \\
Qwen3-VL-30B & \xmark & 46.6 & 38.4 & 35.5 & 39.0 & 39.8 & 35.1 & 35.6 & 38.8 & 47.5 & 68.7 & 52.3 & 55.9 & 67.5 & 61.1 & 57.3 & 57.3 & 57.3 \\
Qwen3-Omni-30B & \cmark & 60.6 & \textbf{62.3} & 48.4 & 35.5 & \textbf{50.9} & 51.0 & \textbf{55.2} & 52.8 & 63.5 & 71.0 & 62.7 & 61.1 & 80.5 & 77.7 & 70.0 & 65.3 & 67.8 \\
\midrule
\rowcolor{rowgray}
\multicolumn{19}{c}{\textbf{Open-source Online-to-Offline VideoLLMs}} \\
\midrule
\textbf{StreamOV} & \cmark & \textbf{64.4} & 56.2 & \textbf{52.4} & \textbf{44.5}  & 48.5  & \textbf{53.6} & 50.4 & \textbf{53.1} &  \textbf{65.6} & \textbf{71.0} & \textbf{64.3}  & \textbf{62.8}  & \textbf{81.2} & \textbf{79.4} & \textbf{70.9} & \textbf{67.3} & \textbf{69.3} \\
\bottomrule
\end{tabular}%
}
\end{table}

\subsection{Evaluation on Visual-Only Benchmarks}

We also conduct experiments on visual-only benchmarks, where each query requires visual understanding without audio-dependent reasoning. In Tab.~\ref{tab:streamingbench}, we report the results of StreamOV on the Visual-Only QA subset of StreamingBench, where our method achieves an improvement of 8.3\% over the Qwen3-Omni baseline. We further report results on another online visual benchmark, OVO-Bench, in Supp.~\ref{supp:ovobench}. 
Please refer to supplementary for more visual-only benchmark results.

\subsection{Ablation Study}

\begin{wraptable}{r}{0.48\columnwidth} 
    \centering
    \caption{Ablation study of StreamOV.}
    \label{tab:ablation}
    \vspace{10pt} 
    \small 
    \begin{tabular}{ccc|c}
        \toprule
        \textbf{Q-Aw.} & \textbf{Q-Ag.} & \textbf{L-Mem.} & \textbf{Avg. Acc} \\
        \midrule
        \xmark & \xmark & \xmark & 73.7 \\
        \cmark & \xmark & \xmark & 80.0 \\
        \xmark & \cmark & \xmark & 78.0 \\
        \cmark & \cmark & \xmark & 80.4 \\
        \cmark & \cmark & \cmark & \textbf{81.6} \\
        \bottomrule
    \end{tabular}
\end{wraptable}

In Tab.~\ref{tab:ablation}, we conduct an ablation study on SOVBench-O subset to evaluate the individual contributions of the three core components in StreamOV: Query-Aware, Query-Agnostic, and Long-Memory. As shown in Table 4, starting from a 73.7\% Avg. baseline without any of these modules, we observe that the inclusion of either Query-Aware or Query-Agnostic features leads to a performance gain, reaching 80.0\% and 78.0\% respectively. Combining both modules further improves the results to 80.4\%, demonstrating their complementary nature in capturing multi-faceted information. Finally, the integration of Long-Memory provides the most significant boost, elevating the overall average to 81.6\%. These results validate that each component is essential and they work synergistically to achieve superior performance in streaming scenarios. More results in Supp.~\ref{supp:add_ablation}.

\section{Conclusion}
In this paper, we introduce a novel framework StreamOV for streaming omni-video understanding task, where models must continuously process synchronized audio-visual streams, maintain bounded long-term memory, and decide when to respond. We further construct two SOVBench variants to evaluate online multi-round audio-visual QA capability and proactive response triggering with intentional silence. StreamOV combines multimodal evidence routing, long-short term memory, and an MLLM-as-a-trigger mechanism, enabling efficient streaming reasoning without explicit silence-token generation or external routers. Experiments on SOVBench-O/T and existing online/offline benchmarks demonstrate its effectiveness among streaming omni-video, offline omni-video and visual-only understanding settings.


\bibliographystyle{unsrt}
\bibliography{ref} 

\include{supp}

\end{document}

%% file: supp.tex
\appendix

\section{Benchmark Analysis}

\begin{table}[t]
\centering
\caption{Comparison with existing video understanding benchmarks. 
SOVBench is designed for streaming omni-video understanding, covering audio-visual perception, multi-turn interaction, proactive response, and intentional silence.}
\label{tab:benchmark_comparison}
\small
\setlength{\tabcolsep}{4pt}
\resizebox{\textwidth}{!}{%
\begin{tabular}{lcccccc}
\toprule
\textbf{Benchmark} 
& {\shortstack{Online /\\Streaming}} 
& {\shortstack{Multi-turn\\Interaction}} 
& {\shortstack{Audio\\Input}} 
& {\shortstack{Proactive\\Response}} 
& {\shortstack{Silence\\Decision}} 
& {\shortstack{Omni-video\\Understanding}} \\
\midrule
Video-MME          & \xmark & \xmark & \cmark & \xmark & \xmark & \xmark \\
MLVU               & \xmark & \xmark & \xmark & \xmark & \xmark & \xmark \\
LongVideoBench     & \xmark & \xmark & \xmark & \xmark & \xmark & \xmark \\
EgoSchema          & \xmark & \xmark & \xmark & \xmark & \xmark & \xmark \\
\midrule
StreamingBench     & \cmark & \xmark & \cmark & \xmark & \xmark & \cmark \\
OVO-Bench          & \cmark & \xmark & \xmark & \cmark & \xmark & \xmark \\
SVBench            & \cmark & \cmark & \xmark & \xmark & \xmark & \xmark \\
\midrule
\textbf{SOVBench (ours)}  & \cmark & \cmark & \cmark & \cmark & \cmark & \cmark \\
\bottomrule
\end{tabular}%
}
\end{table}

Tab.~\ref{tab:benchmark_comparison} highlights the key differences between SOVBench and existing video understanding benchmarks. Most offline benchmarks evaluate video comprehension with fixed input videos and predefined questions, making them insufficient for measuring online interaction ability. Recent streaming benchmarks move toward real-time evaluation, but they still focus on partial capabilities, such as visual-only streaming understanding, timestamp-level response, or multi-turn dialogue, without jointly considering audio input, proactive response, and intentional silence. SOVBench fills this gap by formulating streaming omni-video understanding as a continuous audio-visual interaction task, where models not only answer when sufficient evidence appears, but also remain silent when no response is required.

\section{Benchmark Statistics}\label{supp:bench_statis}

We analyze the temporal interaction structure of SOVBench-O. Among all groups, 694 contain at least one Real-Time turn, 97 contain Recall turns, and 742 contain Proactive turns. These paradigms are not mutually exclusive: 564 groups mix multiple temporal behaviors, with Real-Time + Proactive being the most common combination. Purely single-paradigm groups are also retained, including 182 purely Real-Time groups and 45 purely Recall groups.

In terms of content diversity, SOVBench-O covers 15 top-level categories and 86 fine-grained semantic categories. The source videos span a broad range of domains, including Entertainment, Education, People \& Blogs, Gaming, Science \& Technology, Sports, News \& Politics, and Comedy. At the fine-grained level, the benchmark includes heterogeneous content such as game highlights, documentary profiles, celebrity interviews, event livestreams, short films, music videos, film trailers, parodies, social commentary, and tutorials.

\section{Data Filtering}\label{supp:data_filtering}

As mentioned in Sec.~\ref{sec:data_filter_gen}, to ensure the high quality of the training data, we implement a metadata-driven automated assessment pipeline. As shown in Fig.~\ref{supp:quality_assessor_prompt}, we design a multidimensional scoring rubric that instructs a Large Language Model (LLM) to evaluate each video across five key axes: visual dynamism, narrative coherence, information density, audio-visual alignment, and reasoning value. By analyzing the fine-grained metadata (e.g., ASR transcripts and scene descriptions) rather than the raw video, the assessor efficiently identifies samples with strong cross-modal grounding and complex temporal logic. Only videos achieving a total score of 35 or higher (out of 50) are retained, ensuring that the final dataset is rich in instructional and reasoning value.

\section{Trigger Training Data}\label{supp:triggers_traindata}

Through the data construction pipeline described in Sec.~\ref{sec:data_filter_gen}, we can derive a large number of multi-round dialogues from FineVideo. We use the prompt shown in Fig.~\ref{supp:trigger_pos_prompt} and Fig.~\ref{supp:trigger_neg_prompt} to construct positive and negative samples, where positive samples correspond to moments that require a model response and negative samples correspond to moments where the model should remain silent. 
Notably, the construction logic for negative samples is significantly more stringent than for positive ones, as it incorporates multi-dimensional feasibility filters to exclude ambiguous cases such as persistent visual cues or video-boundary references. This ensures that the model learns to remain silent only when the required information is genuinely absent from both audio and visual streams.
In total, we generated approximately 5,000 video samples, consisting of 2,500 positive and 2,500 negative samples.

\section{Ablation stduy of Triggers}\label{supp:different_triggers}
\begin{table}[]
    \centering
    \caption{Comparison of different trigger architectures under the same experimental setting.}\label{supp:trigger_arch}
    \resizebox{\textwidth}{!}{%
    \begin{tabular}{c|c|ccccc}
    \toprule
    Arch. & Hidden Num &  Precision-1 & Recall-1 & Precision-0 & Recall-0 & F1 \\
    \midrule
    Qwen3omni + Trigger & 1 (last prefilling) & 86.8 & 76.7 & 76.7 & 86.8 & 81.4 \\
    \midrule
    \multirow{2}{*}{StreamOV + Trigger} & 1(last prefilling) & 86.1 & 98.3 & 97.8 & 82.1 & 90.5 \\
     & 2(last prefilling + first decode) & 88.4 & 95.0 & 93.8 & 85.9 & 90.7 \\
    \bottomrule
    \end{tabular}
    }
\end{table}
Table~\ref{supp:trigger_arch} presents the ablation study of our proposed response trigger. We evaluate the binary classification performance for the \textit{Respond} (Class-1) and \textit{Wait} (Class-0) actions. First, compared to the baseline Qwen3omni with our proposed trigger, our StreamOV architecture significantly improves the overall F1 score from 81.4 to 90.5 under the same setting (Hidden Num = 1, using only $h_{t,0}$), demonstrating that StreamOV provides more precise memory and better internal cognitive evidence. Furthermore, increasing the Hidden Num from 1 to 2 in StreamOV yields only a marginal F1 improvement (from 90.5 to 90.7). Since incorporating an additional decoding state ($h_{t,1}$) inherently introduces the computational overhead of auto-regressive generation, relying solely on the prefilling state $h_{t,0}$ (Hidden Num = 1) achieves the optimal trade-off between triggering accuracy and computational efficiency.

\section{Evaluation on Visual-Only Online Benchmark}\label{supp:ovobench}

\begin{table}
\centering
\caption{Evaluation on online Visual-Only QA benchmarks: OVO-bench.}\label{tab:ovobench}
\resizebox{\textwidth}{!}{%
\begin{tabular}{lc|ccccccc|cccc|cccc|c}
\toprule
\multirow{2.5}{*}{\textbf{Model}} & \multirow{2.5}{*}{\textbf{\#Frames}} & \multicolumn{7}{c|}{\textbf{Real-Time Visual Perception}} & \multicolumn{4}{c|}{\textbf{Backward Tracing}} & \multicolumn{4}{c|}{\textbf{Forward Active Responding}} & \multirow{2.5}{*}{\textbf{\shortstack{Overall\\Avg.}}} \\ \cmidrule{3-17}
 &  & OCR & ACR & ATR & STU & FPD & OJR & Avg. & EPM & ASI & HLD & Avg. & REC & SSR & CRR & Avg. &  \\
\midrule
\rowcolor{rowgray}
\multicolumn{18}{c}{\textbf{Open-source Offline VideoLLMs}} \\ \midrule
LLaVA-NeXT-Video-7B & 64 & 69.8 & 59.6 & 66.4 & 50.6 & 72.3 & 61.4 & 63.3 & 51.2 & 64.2 & 9.7 & 41.7 & 34.1 & 67.6 & 60.8 & 54.2 & 53.1 \\
LLaVA-OneVision-7B & 64 & 67.1 & 58.7 & 69.8 & 49.4 & 71.3 & 60.3 & 62.8 & 52.5 & 58.8 & 23.7 & 45.0 & 24.8 & 66.9 & 60.8 & 50.9 & 52.9 \\
InternVL-V2-8B & 64 & 68.5 & 58.7 & 69.0 & 44.9 & 67.3 & 56.0 & 60.7 & 43.1 & 61.5 & 27.4 & 44.0 & 25.8 & 57.6 & 52.9 & 45.4 & 50.1 \\
Qwen2-VL-7B & 64 & 69.1 & 53.2 & 63.8 & 50.6 & 66.3 & 60.9 & 60.7 & 44.4 & 66.9 & 34.4 & 48.6 & 30.1 & 65.7 & 50.8 & 48.9 & 52.7 \\
Qwen2.5-Omni-7B & 64 & 67.8 & 53.2 & 67.2 & 47.2 & 71.3 & 57.6 & 60.7 & 52.2 & 58.1 & 19.9 & 43.4 & 36.3 & 62.8 & 47.5 & 48.9 & 51.0 \\
Qwen3-Omni-30B-A3B & 64 & 79.9 & 61.5 & 75.9 & 54.5 & 77.2 & 64.1 & 68.8 & 57.9 & 67.6 & 37.8 & 55.1 & \textbf{37.7} & 70.0 & 62.9 & \textbf{56.9} & 60.3 \\
\midrule
\rowcolor{rowgray}
\multicolumn{18}{c}{\textbf{Open-source Online Video-LLMs}} \\ \midrule
Flash-VStream-7B & 1fps & 25.5 & 32.1 & 29.3 & 33.7 & 29.7 & 28.8 & 29.9 & 36.4 & 33.8 & 5.9 & 25.4 & 5.4 & 67.3 & 60.0 & 44.2 & 33.2 \\
VideoLLM-online-8B & 2fps & 8.1 & 23.9 & 12.1 & 14.0 & 45.5 & 21.2 & 20.8 & 22.2 & 18.8 & 12.2 & 17.7 & - & - & - & - & - \\
Dispider-7B & 1fps & 57.7 & 49.5 & 62.1 & 44.9 & 61.4 & 51.6 & 54.5 & 48.5 & 55.4 & 4.3 & 36.1 & 18.0 & 37.4 & 48.8 & 34.7 & 41.8 \\
TimeChat-Online-7B & 1fps & 75.2 & 46.8 & 70.7 & 47.8 & 69.3 & 61.4 & 61.9 & 55.9 & 59.5 & 9.7 & 41.7 & 31.6 & 38.5 & 40.0 & 36.7 & 46.7 \\ 
StreamForest-7B & 1fps & 68.5 & 53.2 & 71.6 & 47.8 & 65.4 & 60.9 & 61.2 & 58.9 & 64.9 & 32.3 & 52.0 & 32.8 & 70.6 & 57.1 & 53.5 & 55.6 \\ 
ROMA-7B & 2fps & 63.1 & 53.2 & 68.1 & 39.3 & 69.3 & 58.2 & 58.5 & 55.9 & 47.3 & 23.7 & 42.3 & - & - & - & - & - \\
\midrule
\textbf{StreamOV} & 1fps &  \textbf{95.3} & \textbf{78.0} & \textbf{81.0} & \textbf{66.3} & \textbf{85.2} & \textbf{76.1} & \textbf{79.5}  & \textbf{66.3} & \textbf{75.0} & \textbf{46.2} & \textbf{62.4} & 35.4 & \textbf{74.4} & \textbf{70.0} & 56.4 & \textbf{64.0} \\ 
\bottomrule
\end{tabular}%
}
\end{table}

As shown in Tab.~\ref{tab:ovobench}, compared with Qwen3-Omni-30B, StreamOV improves the performance by 3.7\%; compared with the state-of-the-art online model StreamForest, StreamOV further achieves a gain of 8.4\%. These results demonstrate that our evidence routing strategy is not dominated by audio cues, but can effectively identify, preserve, and utilize visual evidence for online visual understanding.

\section{Evaluation on Long Video Benchmarks}\label{supp:long_bench}

\begin{table}[t]
\centering
\caption{Evaluation on Video-MME. We report our results without using subtitles.}
\label{tab:videomme}
\small
\setlength{\tabcolsep}{5pt}
\begin{tabular}{lc|cc}
\toprule
\multirow{2}{*}{\textbf{Model}} 
& \multirow{2}{*}{\textbf{\#Frames}} 
& \multicolumn{2}{c}{\textbf{Video-MME}} \\ 
\cmidrule(lr){3-4}
& & \textbf{Overall} & \textbf{Long} \\
\midrule
\rowcolor{rowgray}
\multicolumn{4}{c}{\textbf{Open-Source Offline VideoLLMs}} \\
\midrule
LongVA-7B~\cite{zhang2024longva}              & 128  & 52.6 & 46.2 \\
LLaVA-Next-Video-7B    & 32   & 46.6 & --   \\
LLaVA-OneVision-7B     & 32   & 58.2 & --   \\
Kangaroo-7B~\cite{kangaroogroup}            & 64   & 56.0 & 46.6 \\
Qwen2.5-VL-7B          & 1fps & 63.2 & 50.4 \\
Qwen2.5-Omni-7B        & 1fps & 64.3 & --   \\
Qwen3-Omni-30B         & 64 & 68.5 & 58.1   \\
\midrule
\rowcolor{rowgray}
\multicolumn{4}{c}{\textbf{Open-Source Online VideoLLMs}} \\
\midrule
Flash-VStream-7B       & 1fps & 61.2 & 50.3 \\
VideoLLM-online-8B     & 2fps & 52.8 & 44.9 \\
Dispider-7B            & 1fps & 57.2 & --   \\
TimeChat-Online-7B     & 1fps & 62.5 & 49.2 \\
StreamForest-7B        & 1fps & 61.4 & --   \\
\textbf{StreamOV (Ours)} & 64 & \textbf{73.5} & \textbf{63.4} \\
\bottomrule
\end{tabular}
\end{table}

We further evaluate StreamOV on Video-MME under the no-subtitle setting, where no external subtitles are provided and the model must rely on visual perception and audio understanding. 
As shown in Tab.~\ref{tab:videomme}, StreamOV obtains 73.5 overall accuracy and 63.4 accuracy on the long-video subset, achieving the best performance among both offline and online VideoLLMs in our comparison. 
Compared with Qwen3-Omni-30B, StreamOV improves the overall score by 5.0 \%. 
The strong performance on long videos suggests that our compact memory can retain useful historical multi-modal evidence over extended temporal contexts, while preserving the efficiency required for streaming scenarios.

\section{Additional Ablation}\label{supp:add_ablation}

\begin{table}[htbp]
    \centering
    \caption{Detailed ablation study of StreamOV across different subtasks.}
    \label{tab:ablation_detail}
    \begin{tabular}{ccc|ccc}
        \toprule
        \textbf{Q-Aw.} & \textbf{Q-Ag.} & \textbf{L-Mem.} & \textbf{Real-time} & \textbf{Proactive} & \textbf{Recall} \\
        \midrule
        \checkmark & \xmark & \xmark & 87.62 & 76.34 & 57.14 \\
        \xmark & \checkmark & \xmark & 86.67 & 72.52 & 64.29 \\
        \checkmark & \checkmark & \xmark & 86.67 & 77.10 & 64.29 \\
        \midrule
        \checkmark & \checkmark & \checkmark & \textbf{88.57} & \textbf{77.86} & \textbf{64.29} \\
        \bottomrule
    \end{tabular}
\end{table}

To further investigate the contribution of each component, we report the detailed accuracy for Real-time, Proactive, and Recall subtasks. The results in Tab.~\ref{tab:ablation_detail} indicate that each module plays a distinct role: the Q-Aw. module is particularly effective for immediate response (Real-time), while the combination of Q-Ag. and L-Mem. enhances the model's ability to handle long-range dependencies and proactive reasoning. The complete model consistently outperforms all variants, validating the effectiveness of the proposed components.

\section{Case Study}\label{supp:case_study}

\begin{figure}
    \centering
    \includegraphics[width=1.0\linewidth]{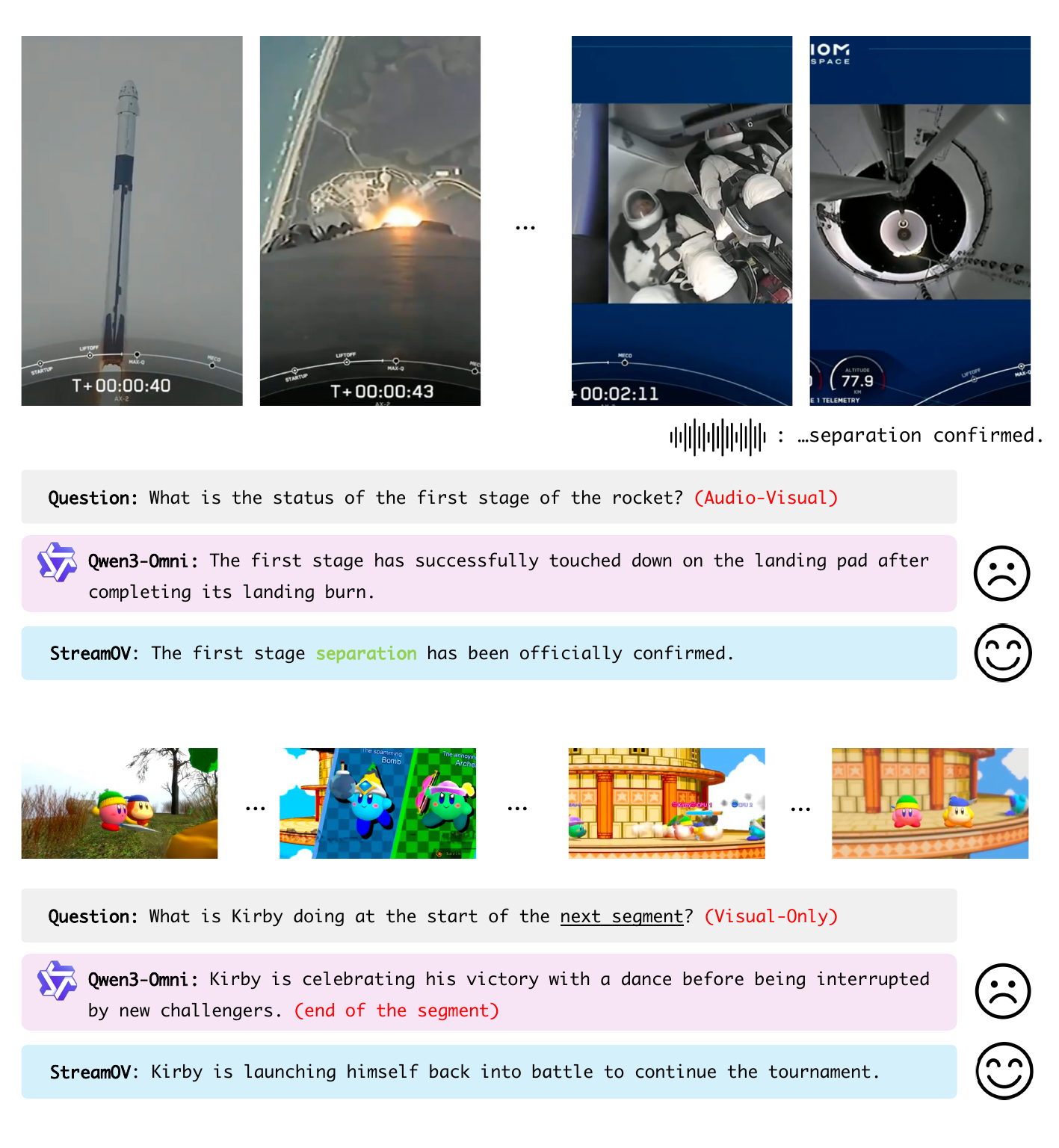}
    \caption{Qualitative cases on streaming audio-visual and visual-only understanding. }
    \label{fig:case1}
\end{figure}

\begin{figure}
    \centering
    \includegraphics[width=\linewidth]{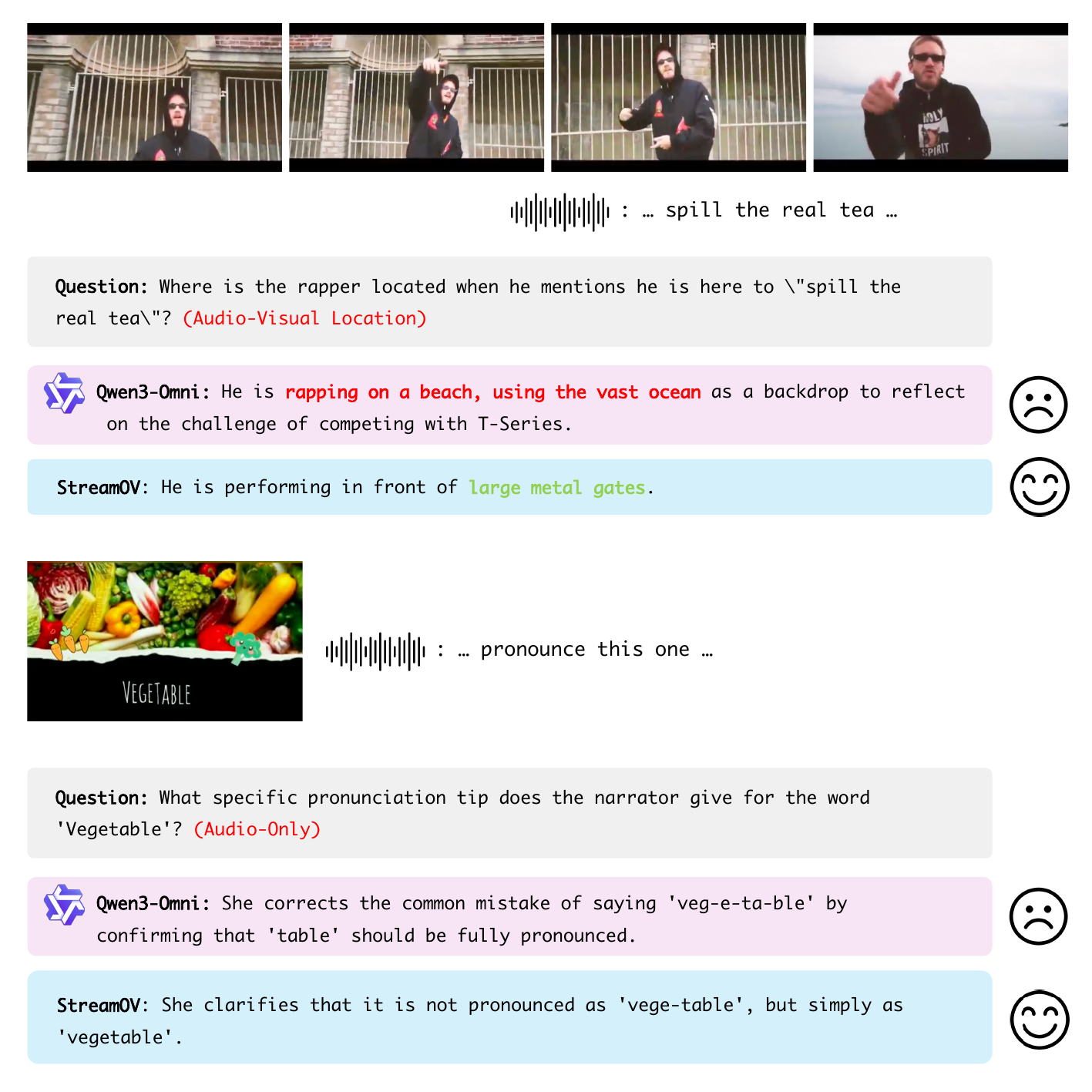}
    \caption{Qualitative cases under different modality requirements.}
    \label{fig:case2}
\end{figure}

As shown in Fig.~\ref{fig:case1} and Fig.~\ref{fig:case2}, StreamOV produces more temporally grounded and modality-aware responses than Qwen3-Omni. In the rocket launch and Kirby examples, Qwen3-Omni answers with temporally mismatched evidence, either hallucinating a later landing event or relying on the previous segment, while StreamOV correctly follows the current audio-visual context and the transition to the next segment. The lower cases further show that StreamOV can adapt to different modality requirements: it grounds the rapper's location in visual evidence while using the audio cue, and accurately captures the audio-only pronunciation instruction. These examples suggest that StreamOV better aligns its responses with the relevant streaming evidence across audio-visual, visual-only, and audio-only scenarios.

\section{Failure Scenarios of Trigger}\label{supp:fail_trigger}

In our experiments, for samples where visual and audio evidence is either sufficiently informative or entirely absent, the model achieves strong performance in determining \textit{when} to respond, benefiting from the inherent reasoning capability of the MLLM and our MLLM-as-a-trigger design. Meanwhile, we also observe that when the audio-visual information is highly sparse, the trigger may become less effective due to the intrinsic limitations of the underlying MLLM.

\section{Limitation and Future Works}\label{supp:limit}
Although StreamOV provides an effective framework for streaming omni-video understanding, several aspects remain worth further exploration. First, our current memory update strategy is still based on predefined evidence scores and heuristic temporal refinement. While this design is efficient and interpretable, future work could develop more adaptive or agentic memory mechanisms that dynamically plan what to observe, store, and retrieve according to evolving user intent. Second, extending the trigger toward more fine-grained actions, such as asking clarification questions, delaying response, or actively seeking additional evidence, would make streaming interaction more flexible. Finally, SOVBench covers diverse online audio-visual scenarios, but it is still a finite benchmark built from curated videos. Scaling it to broader real-world streams with noisier environments and more open-ended user behaviors will further strengthen the evaluation of streaming omni-video systems.

\section{Broader Impact}\label{supp:broader_impact}

This work advances streaming omni-video understanding, which has significant positive potential in real-time applications such as assistive technologies for human-robot interaction. By enabling multi-modal comprehension, it can improve situational awareness in real-world environments. However, like all large-scale multi-modal models, it may inherit biases from training datasets, leading to disparate performance across different demographics. To mitigate such risks, we should explore techniques for enhancing interpretability and controllability of streaming omni-video models in safety contexts.

\section{Disclosure of LLM Usage}\label{supp:llm_usage}

We disclose the use of Large Language Models (LLMs) in the following aspects of this work. (1) Model Architecture: We adopt the open-source \textbf{Qwen3-Omni} as the core backbone of our proposed model for streaming omni-video understanding. (2) Data Processing: LLMs were utilized to process and refine existing open-source datasets. (3) LLMs were used to polish the language, improve grammatical accuracy. All technical contributions and final revisions were conducted and verified by the human authors.

\begin{figure}
\centering
\begin{tcolorbox}[
    colback=gray!5,
    colframe=gray!45,
    width=0.98\linewidth,
    arc=1mm,
    boxrule=0.5pt,
    left=6pt,
    right=6pt,
    top=6pt,
    bottom=6pt,
    title=\textbf{Prompt for Data Quality Assessment},
    coltitle=black,
    colbacktitle=gray!15,
    fonttitle=\small\bfseries
]
\scriptsize 
You are a Data Quality Assessor for multi-modal LLM training. Given \textbf{metadata only}, score the video for training value.

\textbf{\#\# Input JSON (current schema)}
\begin{itemize}[leftmargin=1.2em, noitemsep]
    \item \texttt{video\_uid}: string
    \item \texttt{content}: \{ \texttt{description}, \texttt{qAndA}, \texttt{scenes}: [\{ \texttt{activities}, \texttt{narrative}, \texttt{props}, etc. \}] \}
    \item \texttt{ASR}: list of audio segments as \{ \texttt{end}: seconds, \texttt{text}: string \}.
\end{itemize}

\textbf{\#\# Scoring (each 0-10, integer)}
Use the rubric below. Choose the \textbf{best-matching band} and pick a precise score.

\textbf{1) visual\_dynamism (0-10)}: How visually rich/changing the video is.
\begin{itemize}[leftmargin=1.2em, noitemsep]
    \item \textbf{0-2}: static setup; talking head; minimal motion; no meaningful props.
    \item \textbf{5-6}: multiple actions/scenes; moderate prop interaction.
    \item \textbf{9-10}: highly dynamic; complex interactions; strong visual variety.
\end{itemize}

\textbf{2) narrative\_coherence (0-10)}: Consistency and logical ordering.
\begin{itemize}[leftmargin=1.2em, noitemsep]
    \item \textbf{0-2}: fragmented; contradictory; scenes don't connect; ASR unintelligible.
    \item \textbf{7-8}: clear temporal flow with causes/goals; strong continuity.
\end{itemize}

\textbf{3) information\_density (0-10)}: Amount of specific, non-trivial information.
\begin{itemize}[leftmargin=1.2em, noitemsep]
    \item \textbf{0-2}: mostly filler; few concrete entities; repetitive.
    \item \textbf{9-10}: extremely dense; frequent actionable details; strong training value.
\end{itemize}

\textbf{4) av\_alignment (0-10)}: Alignment between ASR and visual narrative.
\begin{itemize}[leftmargin=1.2em, noitemsep]
    \item \textbf{0-2}: unrelated (generic voiceover vs visuals); frequent mismatches.
    \item \textbf{9-10}: ASR tightly tracks visual actions with clear temporal grounding.
\end{itemize}

\textbf{5) reasoning\_value (0-10)}: Value for audio-visual reasoning (causal/temporal).
\begin{itemize}[leftmargin=1.2em, noitemsep]
    \item \textbf{0-2}: trivial restatement; hard to form non-trivial reasoning QAs.
    \item \textbf{9-10}: dense event graph; abundant grounded entities; supports multi-step reasoning.
\end{itemize}

\textbf{\#\# Task}
Return \textbf{ONLY} a valid JSON object in this schema:
\begin{quote}
\ttfamily\tiny
\{ \\
\quad "video\_uid": "", \\
\quad "scores": \{ "visual\_dynamism": 0, "narrative\_coherence": 0, "information\_density": 0, "av\_alignment": 0, "reasoning\_value": 0, "total\_score": 0 \}, \\
\quad "reasoning": \{ "visual\_analysis": "", "content\_analysis": "", "alignment\_analysis": "", "reasoning\_value\_analysis": "" \}, \\
\quad "recommendation": "Keep" \\
\}
\end{quote}

\textbf{Constraints:} \texttt{total\_score} = sum of five scores. \texttt{recommendation} is "Keep" if \texttt{total\_score} $\ge$ 35, else "Discard".

\textbf{\#\# Input JSON} \\
\{\{PASTE\_RAW\_METADATA\_HERE\}\}
\end{tcolorbox}
\caption{System prompt used for Metadata-based Data Quality Assessment.}
\label{supp:quality_assessor_prompt}
\end{figure}

\begin{figure}[t]
\centering
\begin{tcolorbox}[
    colback=gray!5,
    colframe=gray!45,
    width=0.96\linewidth,
    arc=1mm,
    boxrule=0.5pt,
    left=6pt,
    right=6pt,
    top=6pt,
    bottom=6pt,
    title=\textbf{Prompt for Trigger Sample Construction},
    coltitle=black,
    colbacktitle=gray!15,
    fonttitle=\small\bfseries
]
\small
\textbf{Role:} You are a professional Multimodal Data Engineer. Your task is to extract exactly ONE Positive (Trigger) training sample from a multi-turn video conversation JSON.

\textbf{Rules:}
\begin{enumerate}[leftmargin=1.2em, topsep=2pt, itemsep=0pt]
    \item \textbf{Candidate Selection:}
    \begin{itemize}[leftmargin=1.2em]
        \item Pick ONE QA pair (one user message + the immediately following assistant message) from the SECOND HALF of the "conversation" list.
        \item Skip any messages with category "Proactive".
        \item Any QA with category "Real-time" or "Recall" is acceptable.
    \end{itemize}
    \item \textbf{Construction Details:}
    Let T = the float \texttt{time} of the selected QA pair. Let T\_prev = the \texttt{time} of the most recent ASSISTANT message that occurs strictly BEFORE the selected QA's user message (or 0.0 if none). Keep all float precision to 3 decimal places.
    \begin{itemize}[leftmargin=1.2em]
        \item Assistant content: "\textless Yes\textgreater\ " followed by the original assistant answer text.
        \item Video path: "\{video\_uid\}\_0.00\_\{T\}.mp4"
    \end{itemize}
    \item \textbf{Field Requirements:}
    \begin{itemize}[leftmargin=1.2em]
        \item "summary": A concise natural-language summary of all key information from conversation turns that occur BEFORE the selected QA pair.
        \item "messages": A list with exactly TWO items: user question (prefixed with "\textless video\textgreater\ ") and assistant response ("\textless Yes\textgreater\ " + original answer).
        \item "videos": A list containing one string --- the video path using T.
        \item "t\_prev", "t\_selected": Audit fields (REQUIRED).
    \end{itemize}
    \item \textbf{Output Format (STRICT):} ONE JSON object only. No markdown fences, no explanation.
\end{enumerate}

\textbf{Example Output:}
\begin{quote}
\ttfamily\scriptsize
\{"summary": "The video previously showed the presenter introducing a murder mystery game...", "messages": [\{"role": "user", "content": "\textless video\textgreater\ To recap... "\}, \{"role": "assistant", "content": "\textless Yes\textgreater\ The detective was Kakashi..."\}], "videos": ["0zlxwPIWAqE\_0.00\_54.0.mp4"], "t\_prev": 54.0, "t\_selected": 54.0\}
\end{quote}

\textbf{Please process the following JSON data:}
\end{tcolorbox}
\caption{Prompt for extracting positive trigger samples from raw conversation data.}
\label{supp:trigger_pos_prompt}
\end{figure}

\begin{figure}[t]
\centering
\begin{tcolorbox}[
    colback=gray!5,
    colframe=gray!45,
    width=0.96\linewidth,
    arc=1mm,
    boxrule=0.5pt,
    left=6pt,
    right=6pt,
    top=6pt,
    bottom=6pt,
    title=\textbf{Prompt for Negative (Silence) Sample Construction},
    coltitle=black,
    colbacktitle=gray!15,
    fonttitle=\small\bfseries
]
\small
\textbf{Role:} You are a professional Multimodal Data Engineer. Your task is to extract exactly ONE Negative (Silence) training sample from a multi-turn video conversation JSON, OR explicitly skip if no clean Negative can be constructed.

\textbf{Rules:}
\begin{enumerate}[leftmargin=1.2em, topsep=2pt, itemsep=0pt]
    \item \textbf{Hard Selection Rules:}
    \begin{itemize}[leftmargin=1.2em]
        \item User \texttt{time} MUST be in "valid\_user\_times".
        \item Assistant's category MUST be "Real-time".
        \item NEVER select "Recall" or "Proactive" categories.
    \end{itemize}
    
    \item \textbf{Feasibility Filter:} Reject candidates with these failure modes:
    \begin{itemize}[leftmargin=1.2em]
        \item (F1) Continuous state (e.g., emotional state throughout).
        \item (F2) Video-boundary reference (e.g., "at the start", "close out").
        \item (F3) Omnipresent topic (subject permeates the entire video).
        \item (F4) Persistent visual cue (answer shown by unchanging visual prop).
        \item (F5) "previous-topic" window is too short or empty.
    \end{itemize}

    \item \textbf{Skip Mechanism (REQUIRED):} If no candidate is valid, output: \\
    \texttt{\{"skip": true, "reason": "<short explanation>"\}}

    \item \textbf{Construction Details:}
    \begin{itemize}[leftmargin=1.2em]
        \item Assistant content: exactly "\textless No\textgreater" (no extra text).
        \item Video path: "\{video\_uid\}\_0.00\_\{T\_prev\}.mp4" (T\_prev as placeholder).
        \item Round all float times to 3 decimal places.
    \end{itemize}

    \item \textbf{Field Requirements:} Include "summary", "messages" (user with "\textless video\textgreater\ ", assistant with "\textless No\textgreater"), "videos", "t\_prev", and "t\_selected".
\end{enumerate}

\textbf{Example Output (Sample):}
\begin{quote}
\ttfamily\scriptsize
\{"summary": "The narrator is reciting a poem...", "messages": [\{"role": "user", "content": "\textless video\textgreater\ What does the narrator say about the soil?"\}, \{"role": "assistant", "content": "\textless No\textgreater"\}], "videos": ["9xSyRhe-qOk\_0.00\_51.444.mp4"], "t\_prev": 51.444, "t\_selected": 55.055\}
\end{quote}

\textbf{Example Output (Skip):}
\begin{quote}
\ttfamily\scriptsize
\{"skip": true, "reason": "All candidates reference video boundaries; no clean negative window."\}
\end{quote}

\textbf{Output Format:} ONE JSON object only. No markdown fences, no explanation.

\textbf{Please process the following JSON data:}
\end{tcolorbox}
\caption{Prompt for constructing negative (silence) samples, including strict feasibility filtering and skip mechanisms.}
\label{supp:trigger_neg_prompt}
\end{figure}